\documentclass[10pt,twocolumn,letterpaper]{article}

\usepackage{iccv}
\usepackage{times}
\usepackage{epsfig}
\usepackage{graphicx}
\usepackage{amsmath}
\usepackage{amssymb}
\usepackage{bm}
\usepackage{mathrsfs}
\usepackage{booktabs}
\usepackage{tabularx}
\usepackage[para,online,flushleft]{threeparttable}
\usepackage{multirow}
\usepackage{caption}
\usepackage{subcaption}
\usepackage[ruled,linesnumbered]{algorithm2e}
\usepackage{color, xcolor}
\usepackage[title]{appendix}
\usepackage{dsfont}
\usepackage{mathalias}
\usepackage{pifont}
\definecolor{citecolor}{RGB}{119,185,0}

\usepackage[pagebackref=true,breaklinks=true,letterpaper=true,colorlinks,citecolor=citecolor,bookmarks=false]{hyperref}

\iccvfinalcopy 

\def\iccvPaperID{3938} 

\usepackage{color}
\usepackage{bm}

\DeclareMathOperator{\Tr}{Tr}

\definecolor{azure(colorwheel)}{rgb}{0.0, 0.5, 1.0}

\begin{document}

\title{A Unified Objective for Novel Class Discovery}

\author{Enrico Fini$^{\textcolor{azure(colorwheel)}{1}}$\quad Enver Sangineto$^{\textcolor{azure(colorwheel)}{1}}$\quad St\'{e}phane Lathuili\`{e}re$^{\textcolor{azure(colorwheel)}{2}}$\quad Zhun Zhong$^{\textcolor{azure(colorwheel)}{1}}\thanks{Corresponding author}$\quad Moin Nabi$^{\textcolor{azure(colorwheel)}{3}}$\quad Elisa Ricci$^{\textcolor{azure(colorwheel)}{1,4}}$ \\
\normalsize{$^{\textcolor{azure(colorwheel)}{1}}$~University of Trento, Trento, Italy\quad  $^{\textcolor{azure(colorwheel)}{2}}$~LTCI, T{\'e}l{\'e}com Paris, Institut Polytechnique de Paris, France}\\ \normalsize{$^{\textcolor{azure(colorwheel)}{3}}$~SAP AI Research, Berlin, Germany\quad $^{\textcolor{azure(colorwheel)}{4}}$~Fondazione Bruno Kessler, Trento, Italy} \\
}

\maketitle

\begin{abstract}
In this paper, we study the problem of Novel Class Discovery (NCD). NCD aims at inferring novel object categories in an unlabeled set by leveraging from prior knowledge of a labeled set containing different, but related classes. Existing approaches tackle this problem by considering multiple objective functions, usually involving specialized loss terms for the labeled and the unlabeled samples respectively, and often requiring auxiliary regularization terms. In this paper we depart from this traditional scheme and introduce a UNified Objective function (UNO) for discovering novel classes, with the explicit purpose of favoring synergy between supervised and unsupervised learning. Using a multi-view self-labeling strategy, we generate pseudo-labels that can be treated homogeneously with ground truth labels. This leads to a single classification objective operating on both known and unknown classes. Despite its simplicity, UNO outperforms the state of the art by a significant margin on several benchmarks ($\approx$+$10\%$ on CIFAR-100 and +$8\%$ on ImageNet). The project page is available at : \url{https://ncd-uno.github.io}. 

\end{abstract}

\section{Introduction}
\label{Introduction}

Deep learning has enabled astounding progresses in computer vision.
However, the necessity of large annotated training sets for these models is often a limiting factor. For instance, training a deep neural network for classification requires a large amount of labeled data for each class of interest. This constraint is even more severe in scenarios where collecting sufficient data for each class is expensive or even impossible, as for instance in medical applications.

\vspace{-.06in}
To alleviate these problems, {\em Novel Class Discovery (NCD)} \cite{han2020automatically,han2021autonovel,han2019learning} has recently emerged as a practical solution.
NCD aims at training a network that can simultaneously classify a set of labeled classes while discovering new ones in an unlabeled image set. The basic motivation behind this setting is that the network can benefit from the supervision available on the labeled set to learn rich image representations that can be transferred to discover unknown classes in the unlabeled set. At training time, data are split into a set of labeled images and a set of unlabeled images, assuming disjoint sets of classes. These two sets are used to train a single network to classify both the known and the unknown classes.
\begin{figure}
    \centering
    \includegraphics[width=0.95\columnwidth]{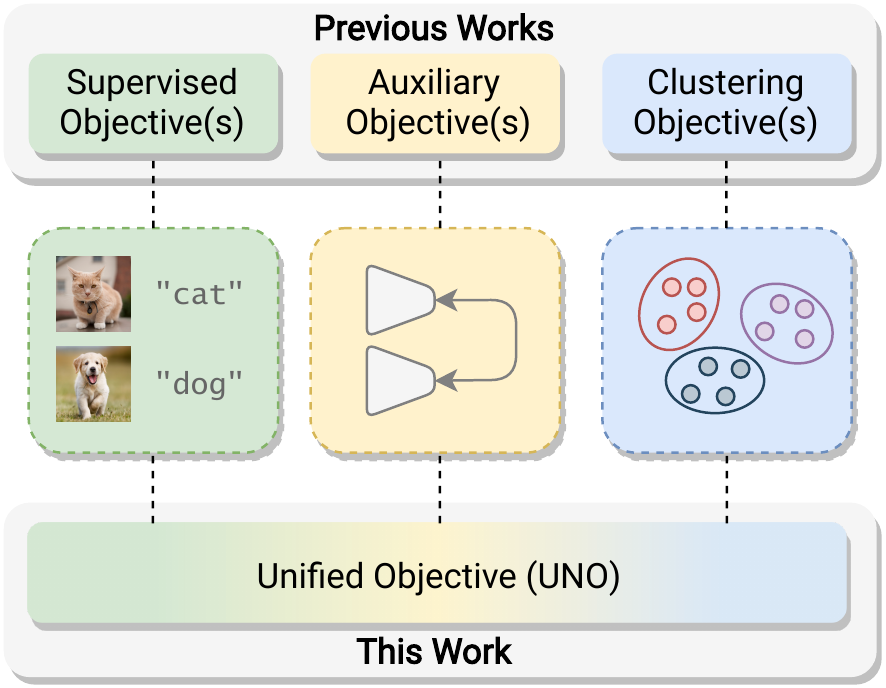}
    \vspace{-.1in}
    \caption{A visual comparison of our \textit{UNified Objective (UNO)} with previous works. Existing approaches tackle NCD using multiple objective functions such as \textit{supervised}, \textit{clustering} and \textit{auxiliary} objectives. On the contrary, we devise a single classification objective operating on both known and unknown classes.}
    \vspace{-.08in}
    \label{fig:teaser}
\end{figure}
Note that this problem is similar but different from {\em semi-supervised learning} \cite{sohn2020fixmatch,van2020survey}, because, in the latter, the working assumption 
is that labeled and unlabeled sets share the same classes. Differently, in NCD, the two sets of classes are supposed to be disjoint. Moreover, differently from common {\em clustering}~\cite{caron2018deep,van2020scan} scenarios, in an NCD framework, labeled data 
can be utilized at training time, and the challenge consists in transferring the supervised knowledge on the known classes to improve clustering of the unknown ones.

\vspace{-.07in}
Most NCD methods usually perform an initial \emph{supervised pretraining} step on the labeled set, followed by a \emph{clustering} step on the unlabeled data  \cite{han2019learning,hsu2017learning,hsu2019multi}. This simple pipeline provides an effective means to transfer the representation capability from the labeled set to the unlabeled one. Generally speaking, these approaches combine two separated objectives. On the one hand, there is direct supervision through labels on the labeled set. 
On the other hand, a clustering objective is used to discover the novel categories. 
Clustering objectives are generally based on pseudo-labels~\cite{han2020automatically,iscen2019label,jia21joint,zhao2021novel,Zhong_2021_CVPR,zhong2021openmix} estimated on the unlabeled set. In practice, these objectives are combined through independent losses such as cross-entropy (CE) and binary cross-entropy (BCE), respectively. Usually, the BCE loss is computed with pseudo pairwise labels often determined by setting an ad-hoc threshold which heavily influences the performance of these methods. 

\vspace{-.07in}
Additionally, NCD approaches generally require a strong semantic similarity between labeled and unlabeled classes in order to obtain expressive representation for discovering new concepts.
In order to decrease the bias of the features toward known classes, Han \etal \cite{han2020automatically} propose to use an additional phase of \emph{self-supervised pretraining} on \emph{all} available images, both labeled and unlabeled,  before the \emph{supervised pretrain}. Moreover, the clustering stage is strengthened with another self-supervised objective (consistency), which enforces the model to output similar predictions for two different data augmentations of the same image. Adding an additional auxiliary objective makes optimization of this model even more cumbersome as it requires to further tune the hyper-parameters for each of these competing objectives. Moreover, this method assumes the availability of the unlabeled set in the pretraining stage. This is not suitable when it comes to learning in a sequential fashion as it requires to repeat the costly self-supervised pretraining stage every time that the unlabeled set changes. 

\vspace{-.05in}
Motivated by the need of simplifying NCD approaches, and inspired by the recent advancements in self-supervised learning \cite{caron_swav, simclr}, in this paper we propose to eliminate the self-supervised pretraining step and unify all the objectives through a single loss function (see Fig.~\ref{fig:teaser}). Specifically, using  a  multi-view self-labeling strategy, 
we generate pseudo-labels that can be treated homogeneously with ground truth labels. This makes it possible to use a unified cross-entropy loss on both the labeled and the unlabeled set. In more detail, given a batch of images, we generate two views of each image using random transformations. 
Then, our network predicts a probability distribution over all classes (\emph{labeled} + \emph{unlabeled}) for each view.
This results in two sub-batches that are independently clustered, so the cluster-assignment of each view is simply used as the pseudo-label for the other view.  Ground truth and pseudo-labels are then used in combination to provide feedback to the network and update its parameters.  
Importantly, using a unified framework that operates on the complete class set enables us to learn a single model that can jointly recognize both labeled and unlabeled classes. We emphasize that this is a key point that is often neglected in the existing solutions for the NCD task. 

\noindent \textbf{Contributions.} Our contributions can be summarized as follows: \textbf{(i)} we introduce a UNified Objective (UNO) for NCD where cluster pseudo-labels are treated homogeneously with ground truth labels, allowing a single CE loss to operate on both labeled and unlabeled sets; 
\textbf{(ii)} using multi-view, multi-head and over-clustering strategies we learn powerful representations while discovering new classes, de facto eliminating the need for self-supervised pretraining in NCD;
\textbf{(iii)} experimentally, we show that our method surpasses all previous works on three publicly available benchmarks by a large margin. Notably, we outperform previous methods by $8\%$ in accuracy on ImageNet, and by $\approx$+$10\%$ on CIFAR-100. 
\textbf{(iv)} Finally, we push NCD to the limit by changing the proportion of labeled and unlabeled samples, and find that our objective outperforms the state-of-the-art even more significantly on complex benchmarks.

\section{Related Work}
\label{Related}

\noindent\textbf{Novel Class Discovery.} The concept of Novel Class Discovery was first formally introduced by Han \textit{et al.} in~\cite{han2019learning}, but the study of NCD can be dated back to the works  in~\cite{hsu2017learning,hsu2019multi}. In \cite{hsu2017learning}, Hsu \etal introduce the problem of transferring clustering models over tasks, which corresponds to the NCD setting: the goal is to cluster an unlabeled dataset given a labeled dataset without class-overlap. In \cite{hsu2017learning,hsu2019multi}, a prediction network is trained on labeled data, which is then used to estimate pairwise similarities between unlabeled samples. Finally, the clustering network is trained to recognize novel classes in unlabeled dataset by using the predicted pairwise similarities. The main difference between \cite{hsu2017learning} and \cite{hsu2019multi} lies in the choice of the training loss applied to the prediction network.

\vspace{-.05in}
More recently, Han \etal~\cite{han2019learning} address the same problem proceeding in two steps: a data embedding is learned on the labeled data using a metric learning technique, and then fine-tuned while learning the cluster assignments on the unlabeled data. Interestingly, they also tackle the problem of estimating the number of classes in the unlabeled dataset.
Latter, many NCD works~\cite{han2020automatically,jia21joint,zhao2021novel,Zhong_2021_CVPR,zhong2021openmix} are designed following the two-step training strategy. Han \etal~\cite{han2020automatically} find that pretraining the backbone network in a self-supervised manner using rotation prediction can significantly improve the clustering accuracy. In addition, they employ rank statistics to identify data pairs that belong to the same class and minimize BCE to bring the network output closer for these pairs similarly to~\cite{hsu2019multi}.
This pseudo-labeling loss is minimized together with a CE loss on the labeled set and a consistency loss that enforces network invariance to some random data transformations. OpenMix~\cite{zhong2021openmix} generates virtual samples by mixing the labeled and unlabeled data, which can resist the noisy labels of unlabeled data. To leverage more positive samples, Zhong \etal~\cite{Zhong_2021_CVPR} introduce Neighborhood Contrastive Learning (NCL) to aggregate pseudo-positive
pairs with contrastive learning.

\vspace{-.05in}
From this literature review, we observe that existing methods commonly need to (i) learn the classifier of the novel classes with pairwise relations between unlabeled samples, (ii) use a consistency loss to enforce network invariance to data transformations, and (iii) jointly train the network with several losses. Different from them, in this work, we propose a unified framework which enforces data-transformation consistency via a pseudo-labeling process by using a single objective. 

\noindent\textbf{Deep Clustering.}
Identifying classes in an unsupervised manner can be formalized as a clustering problem. {Deep Cluster}~\cite{caron2018deep} can be considered as the first clustering method capable of learning rich image representations with deep networks without supervision. This approach alternates between a k-mean step that provides pseudo-labels and a network training step where cluster assignments are used as supervision. 
More recently, Van Gansbeke \etal~\cite{van2020scan} also show that a two-step approach where feature learning and clustering are decoupled can lead to state-of-the-art performance. 
Several approaches have been proposed to avoid this iterative process.
The training stability of {Deep Cluster} is improved in \cite{zhan2020online} thanks to an online training formulation. A deep clustering network is trained in an end-to-end manner in \cite{ji2019invariant} and \cite{menapace2020learning} thanks to a mutual-information maximization objective. 
Other approaches propose more sophisticated pseudo-labeling strategies: Asano \etal \cite{YM.2020Self-labelling} employ an optimal transport formulation to obtain robust pseudo-labels. This formulation is based on the Sinkhorn-Knopp algorithm \cite{cuturi2013sinkhorn} that maps sample representations to prototypes. Caron \etal \cite{caron_swav} propose to use this clustering algorithm to introduce a "swapped" prediction mechanism that uses two random transformations of the same images, referred to as views. Cluster assignments are estimated for each view and are used as a pseudo-label for the other view. In this work, we take advantage of this swapped prediction mechanism to obtain pseudo-labels for the unlabeled set but we incorporate to this mechanism the network-head corresponding to the labeled classes to guide clustering.

\section{Method}
\label{sec:method}

In the NCD task, training data are split in two sets: a labeled set $D^l\!=\!\{(\xmat_1^l,y_1^l), ..., (\xmat_N^l,y_N^l)\}$ and an unlabeled set $D^u\!=\!\{\xmat_1^u, ..., \xmat_M^u\}$, where each $\xmat_i^l$ in $D^l$ and $\xmat_j^u$ in $D^u$ is an image and $y_i^l \in \mathcal{Y}^l\!=\!\{1, ..., C^l\}$ is a categorical label. The one-hot representation of $y_i^l$ is denoted as $\yvect_i^l$. The goal is to use $D^u$ to discover $C^u$ clusters, where  $C^u$ is known a priori. The set of $C^l$ labeled classes is assumed to be disjoint from the set of $C^u$ unlabeled classes.
Note that, at test time, we aim at classifying images corresponding to both labeled and unlabeled classes.
We formulate this problem as learning a mapping from the image domain to the complete-label set 
$\mathcal{Y} = \{1, ..., C^l, C^l\!+\!1, ..., C^l\!+\!C^u  \}$, where the first $C^l$ elements correspond to $\mathcal{Y}^l$, while the subsequent $C^u$ elements correspond to latent classes which should emerge from the clustering process. 

In the following subsections, we first introduce how our Unified Objective learns this mapping (Sec.~\ref{sec:uno}), then we explain how to use a multi-view self-labeling strategy to obtain strong pseudo-labels (Sec.~\ref{sec:pseudolabel}), and finally we show how the performance of our method can be boosted by using multi-head clustering and overclustering (Sec.~\ref{sec:details}).

\begin{figure*}[t]
    \centering
    \includegraphics[width=0.95\textwidth]{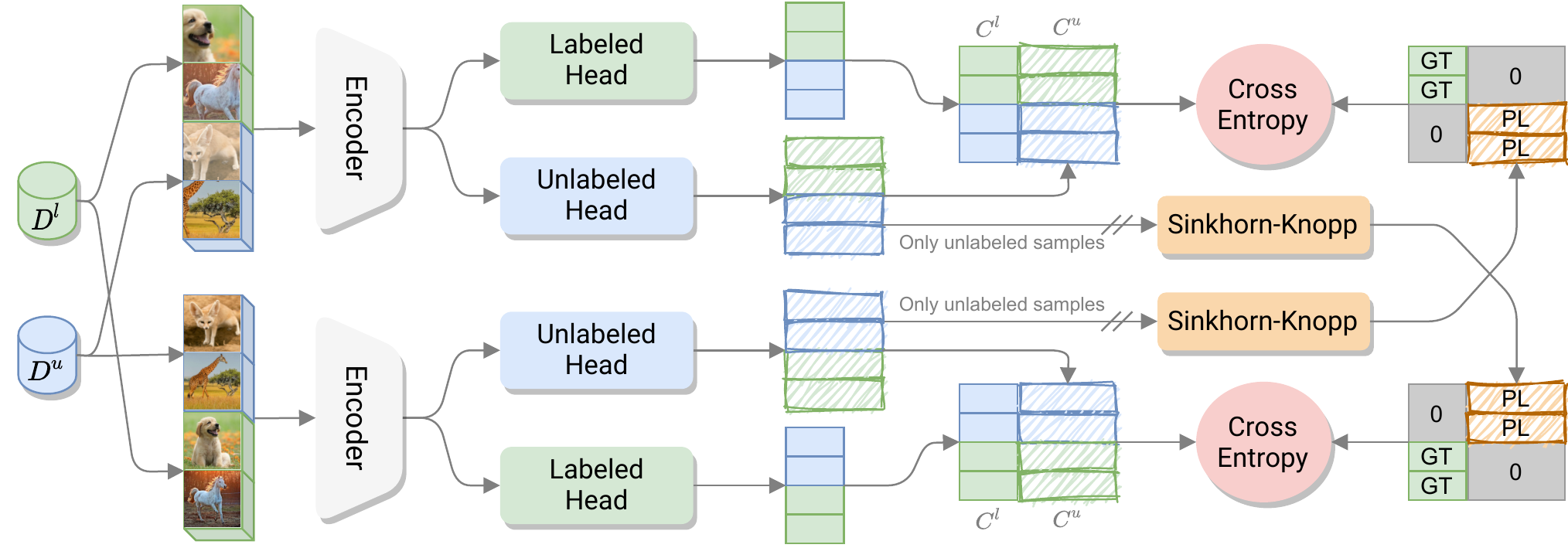}
    \vspace{-.1in}
    \caption{Overview of the proposed architecture. In green we represent the ``labeled components'' (labeled subset $D^l$, labeled head $h$, labeled samples), in blue their unlabeled counterparts (unlabeled subset $D^u$, unlabeled head $g$, unlabeled samples) and in orange the pseudo-labeling algorithm and its outputs. Sketchiness indicates uncertainty in the unlabeled logits and pseudo-labels. The parameters of the encoder $E$ and the heads ($h$ and $g$) are shared for the two views. 
    }
    \label{fig:method}
\end{figure*}

\subsection{Unified Objective}
\label{sec:uno}

To solve the NCD problem, we propose to train a neural network $f_{\theta}$, parametrized by $\theta$, which computes the posterior probabilities over $\mathcal{Y}$: $f_\theta(\xmat)\!=\!\{p(y|\xmat); \!y \in \mathcal{Y}$\}.
Our network architecture is shown in Fig.~\ref{fig:method}: it is composed of a shared encoder $E$ and two heads, $h$ and $g$. 
The encoder $E$ is a standard convolutional network (CNN) followed by an average pooling layer, and $\bm{z}\!=\!E(\xmat)$, $\zvect\in\mathbb{R}^k$, is a feature vector representing the input image $\xmat$. The first head $h$ is a linear classifier with $C^l$ output neurons. On the other hand, $g$ is implemented using a Multilayer Perceptron (MLP), that projects $\zvect$ to a lower dimensional representation $\zvect'$, and a linear classifier with $C^u$ output neurons. Following \cite{caron_swav, wang2020understanding}, we $l2$-normalize $\zvect$, $\zvect'$ and the linear classifiers. 

Importantly, the logits $\bm{l}_h\in\mathbb{R}^{C^l}$ and $\bm{l}_g\in\mathbb{R}^{C^u}$ respectively produced by $h$ and $g$ are concatenated: $\bm{l}\!=\![\bm{l}_h, \bm{l}_g]$. Then, they are fed to a shared softmax layer $\sigma$ which outputs a posterior distribution  over the complete-label set 
$\mathcal{Y}$: $\pvect\!=\!\sigma(\lvect / \tau)$, where $\tau$ is the temperature of the softmax. Once we have $\bm{p}$, we can train the whole network $f$ using standard cross-entropy:
\begin{equation}
\label{eq.cross-entropy}    
\ell(\xmat,\yvect) = - \sum_{c=1}^C \yvect_c \log\left(\pvect_c\right),
\end{equation}
\noindent
where $C\!=\!C^l\!+\!C^u$. $\yvect_c$ and $\bm{p}_c$ are the $c$-th elements of the label $\yvect$ and the network prediction $\bm{p}\!=\!f(\xmat)$, respectively.
The label $\yvect$ used for image $\xmat$ depends on whether $\xmat \in D^l$ or $\xmat \in D^u$. If $\xmat$ belongs to the labeled dataset we apply zero-padding to $\yvect^l$, while if $\xmat \in D^u$, we zero-pad the pseudo-label $\hat{\yvect}$ associated with $\xmat$: 
\begin{equation}
\label{eq:pseudolabel}
  \yvect =
    \begin{cases}
      [\yvect^l{}, \bm{0}_{C^u} ] & \xmat \in D^l\\
      [\bm{0}_{C^l}, \hat{\yvect} ]  & \xmat \in D^u.
    \end{cases}       
\end{equation}
Here, $\bm{0}_{C^u}$ and $\bm{0}_{C^l}$ denote zero vectors of dimension $C^u$ and $C^l$, respectively. This padding formulation is a natural choice, which derives from the assumption that the known and unknown classes are disjoint.

\subsection{Multi-view and Pseudo-labeling}
\label{sec:pseudolabel}
In this section, we show how a multi-view strategy can be  leveraged to generate pseudo-labels for our Unified Objective. Given an image $\xmat$, we adopt common data-augmentation techniques, consisting in applying random cropping and color jittering to $\xmat$, and we obtain two different ``views'' of $\xmat$, which are resized to the original size and fed to $f$.  These data-augmentation techniques, originally exploited in the self-supervised learning field \cite{simclr}, have recently been successfully applied also to  standard supervised learning \cite{khosla2020supervised}. Coherently, we extract two views $\vmat_1$ and $\vmat_2$ from $\xmat$, both when $\xmat \in D^l$ and when $\xmat \in D^u$ (see Fig.~\ref{fig:method}).

In case $(\xmat, \yvect^l) \in D^l$, we associate $\vvect_1$ and $\vvect_2$ with the same label $\yvect_1 = \yvect_2 = [\yvect^l, \bm{0}_{C^u} ]$.
On the other hand, if $\xmat \in D^u$, then we use $\vvect_1$ to compute $\hat{\yvect}_1$ and $\vvect_2$ to compute $\hat{\yvect}_2$ and then we plug both pseudo-labels in Eq.~\eqref{eq:pseudolabel}. At this point, Eq.~\eqref{eq.cross-entropy} can be applied independently for each view. However, this approach does not encourage the network to output consistent predictions for different views of the same image. In order to enforce such behavior, following~\cite{caron_swav}, we use the {\em swapped prediction task}:
\begin{equation}
    \ell(\vvect_1,\yvect_2) + \ell(\vvect_2,\yvect_1).
\end{equation}
When we evaluate each term in the above formula, we apply a ``stop-gradient'' for the pseudo-label, \textit{i.e.}, the gradient flows only though $f(\vvect_1)$. Note that these two loss terms are instances of the same objective applied to different views.

Regarding the computation of pseudo-labels, a na\"ive solution to obtain $\hat{\yvect}_1$ given $\vvect_1$, would be to simply use the predictions of $g(\zvect_1)$, with $\zvect_1 = E(\vmat_1)$. 
Let  $\pvect_g^1 = \sigma(\lvect_g^1 / \tau)$, 
where $\lvect_g^1$ are the logits computed by $g(\bm{z}_1)$ and the softmax operation is applied to {\em only}
the $C^u$ output neurons of $g(\zvect_1)$. We may set $\hat{\yvect}_2 = \pvect_g^1$ and use $\hat{\yvect}_2$ in Eq.~\eqref{eq:pseudolabel} to get $\yvect_2$.
However, as observed in \cite{YM.2020Self-labelling}, this pseudo-label assignment may lead to degenerate solutions, in which, \eg, $g$ always predicts the same logits vector, for any input. In this case, in Eq.~\eqref{eq.cross-entropy}, the network prediction and the labels are basically the same and there is no learning. Instead, inspired by \cite{caron_swav,YM.2020Self-labelling}, when computing $\hat{\yvect}_2$, we
add an entropy term which penalizes those situations in which all the logits are equal to each other and incentives a uniform partition of the pseudo-labels over all the  $C^u$ clusters. Specifically, let $\bm{L} = [\bm{l}_g^1, ..., \bm{l}_g^B]$ be the matrix whose columns are the logits computed by $g$ with respect to a mini-batch of images of size $B$. Moreover, let $\hat{\Ymat} = [\hat{\yvect}_1, ..., \hat{\yvect}_B]^\top$ be the matrix whose rows are the unknown pseudo-labels of the current batch. $\hat{\Ymat}$ is found by solving:
\begin{equation}
\label{eq.Swav-optimization} 
\hat{\Ymat} = \max_{\Ymat \in \Gamma}  \Tr (\Ymat \bm{L}) + \epsilon \operatorname{H}(\Ymat),
\end{equation}
\noindent
where $\epsilon\! >\! 0$ is an hyper-parameter, $\operatorname{H}$ is the entropy function which is used to ``scatter'' the pseudo-labels, $\Tr$ is the trace function, and $\Gamma$ 
is the transportation polytope defined as:
\begin{equation}
\small
\label{eq.polytope} 
 \Gamma = \{ \Ymat \in \mathbb{R}^{C^u \times B}_+ | \Ymat  \bm{1}_B = \frac{1}{C^u} \bm{1}_{C^u}, 
 \Ymat^\top  \bm{1}_{C^u} = \frac{1}{B} \bm{1}_{B} \}.
\end{equation}

These constraints enforce that, on average, each
cluster is selected $\frac{B^u}{C^u}$ times in the batch, where $B^u$ is the number of unlabeled samples in the batch.
The solution to Eq.~\eqref{eq.Swav-optimization}
is obtained using the Sinkhorn-Knopp
algorithm \cite{cuturi2013sinkhorn} (for more details, we refer to \cite{YM.2020Self-labelling}). 
The resulting pseudo-labels, represented by each row $\hat{\yvect}_i$ in $\hat{\Ymat}$ can then be discretised. However, we found that the best performance can be achieved using soft pseudo-labels $\hat{\yvect}_i \in [0,1]^{C^u}$.

\subsection{Multi-head Clustering and Overclustering}
\label{sec:details}
In order to boost the clustering performance, inspired by~\cite{ji2019invariant}, jointly with the main clustering task, we also adopt overclustering, \ie, we force $f$ to produce an alternative partition of the unlabeled data which is more fine-grained. This is known to enhance the quality of the representations. Specifically, an overclustering head $o$, connected with $E$, is similar to $g$, but with $K = C^u \times m$ cluster output neurons. 

Additionally, inspired by \cite{caron2018deep,ji2019invariant}, we also use multiple clustering ($g_1, ..., g_n$) and overclustering ($o_1, ..., o_n$) heads. This is useful because heads could converge to suboptimal clustering configurations. By using multiple heads we can smooth down this effect and increase the overall signal backpropagated to the shared part of the network.
At training time, for a given batch of data, we iterate over $g_1, ..., g_n$ and, for each head $g_i$, 
we concatenate the logits produced by $h$ ($\bm{l}_h$) with the logits produced by $g_i$ (\ie $\bm{l}_{g_i}$). We feed the result to the $C^l+C^u$-element softmax layer and, following the procedure described above, we compute Eq.~\eqref{eq.cross-entropy} for each $\xmat$ in the batch. Similarly, for each $o_j$, we concatenate $\bm{l}_h$ with the logits produced by $o_j$ (\ie $\bm{l}_{o_j}$), we use a $C^l + K$-element softmax layer and we again compute Eq.~\eqref{eq.cross-entropy} for each $\xmat$ in the batch.

\section{Experiments}
\label{sec:experiments}
\subsection{Experimental Setup}
\label{sec:setup}
\noindent \textbf{Datasets.} We evaluate the performance of our method on three well-established NCD benchmarks following \cite{han2019learning, han2020automatically}, \textit{i.e.}, CIFAR10~\cite{krizhevsky2009learning}, CIFAR100~\cite{krizhevsky2009learning} and ImageNet~\cite{deng2009imagenet}.
Each dataset is divided into two subsets, the labeled set that contains labeled images belonging to a set of known classes, and an unlabeled set of novel classes for which we do not possess any supervision except for the number of classes. 
The details of the splits are shown in Tab.~\ref{tab:datasets}.
Standard data splits used in the literature (1, 2 and 4 in Tab.~\ref{tab:datasets}) exhibit either (i) a \textit{small number of classes} or (ii) a \textit{strong imbalance} in the number of classes of the two subsets (the labeled set is much larger than the unlabeled set). However, these two assumptions do not hold in real-world scenarios, where the unlabeled data is far more abundant than its labeled counterpart. Hence, we introduce a new split that better approximates practical applications for NCD: CIFAR100-50. As shown in Tab.~\ref{tab:datasets}, CIFAR100-50 contains a large number of unlabeled classes (50), making the task more challenging. In our experiments, we show that the performance of all methods in the proposed split drops considerably in comparison to the easier one (CIFAR100-20). This gives evidence that current solutions for NCD are not yet ready for deployment. Even more challenging settings are analyzed in Fig.~\ref{fig:accu_num_unlab}.

\begin{table}[t]
  \small
  \centering
  \resizebox{0.9\columnwidth}{!}{\begin{tabular}{lcccc}
    \toprule
    \multirow{2}[3]{*}{Dataset}  &
    \multicolumn{2}{c}{Labeled} & \multicolumn{2}{c}{Unlabeled}\\
    \cmidrule(lr){2-3} \cmidrule(lr){4-5}
     & Images & Classes & Images & Classes \\
    \midrule
    (1) CIFAR10 & 25K & 5 & 25K & 5 \\
    (2) CIFAR100-20 & 40K & 80 & 10K & 20 \\
    (3) CIFAR100-50 & 25K & 50 & 25K & 50 \\
    (4) ImageNet & 1.25M & 882 & $\approx$30K & 30\\
    \bottomrule
  \end{tabular}}
  \vspace{-.1in}
  \caption{Statistics of the datasets and splits used in our novel class discovery benchmark.}
  \label{tab:datasets}
\end{table}

\noindent\textbf{Protocol.} We evaluate our model using two evaluation settings: \textbf{task-aware} and \textbf{task-agnostic}. In the task-aware evaluation, we use the task information to exclude those outputs that are not relevant for the current sample. In other words, given an image belonging to the labeled classes we only consider the labeled classifier, and viceversa for images belonging to unlabeled classes we only consider the output of the unlabeled heads. This evaluation is typically used in the literature. However, in practical scenarios this evaluation is not very meaningful because it does not assess if the model is able to discriminate labeled from unlabeled classes. Hence, we also report task-agnostic accuracy, where the prediction is simply the most likely output after concatenating labeled and unlabeled logits. 
For the task-aware protocol, we report performance on the training set (Tab.~\ref{tab:state-of-the-art}) and test set (Tab.~\ref{tab:ablation}), while for the task-agnostic protocol and  we report performance on the test set (Tab.~\ref{tab:ablation} and Tab.~\ref{tab:concat_eval}).
The results in Tab.~\ref{tab:ablation} are averaged over 3 runs, while for Tab.~\ref{tab:state-of-the-art} and~\ref{tab:ablation} are averaged over 10 runs following the protocol of~\cite{han2020automatically}. 

\noindent\textbf{Metrics.} We use the accuracy measure for labeled samples and the average clustering accuracy for unlabeled samples. The average clustering accuracy is defined as:
\begin{equation}
\operatorname{ClusterAcc} = \max _{perm \in P}
\frac{1}{N} \sum_{i=1}^{N}
\mathds{1}
\left \{
  {y}_{i}=perm\left(\hat{y}_{i}\right)
\right \},
\end{equation}
where $y_i$ and $\hat{y}_{i}$ represent the ground-truth label and predicted label of a sample $\xmat_{i} \in D^u$, respectively. $P$ is the set of all permutations, which is computed with the Hungarian algorithm~\cite{kuhn1955hungarian}.
Since we train the network using multiple heads, we compute evaluation metrics independently for each head and report both average accuracy and best head accuracy. We define the best head as the one exhibiting lowest training loss in the last epoch.

\begin{table*}[!t]
  \centering
  \small
  \begin{tabular}{lccccccccccccccc}
  \toprule
    \multirow{4}{*}{Method} &
    \multirow{4}{*}{Concat}  &
    \multirow{4}{*}{Over} &
    \multirow{4}{*}{Aug} &
    \multicolumn{6}{c}{CIFAR10} & \multicolumn{6}{c}{CIFAR100-50} \\
    \cmidrule(lr){5-10} \cmidrule(lr){11-16}
    &&&& \multicolumn{3}{c}{Task-aware} &  \multicolumn{3}{c}{Task-agnostic} & \multicolumn{3}{c}{Task-aware} &
    \multicolumn{3}{c}{Task-agnostic} \\
    \cmidrule(lr){5-7} \cmidrule(lr){8-10} \cmidrule(lr){11-13} \cmidrule(lr){14-16} 
    &&&& Lab & Unlab & All & Lab & Unlab & All & Lab & Unlab & All & Lab & Unlab & All\\
 \midrule
 \multirow{4}{*}{UNO}
  & \ding{55} & \ding{51} & Strong & 90.6 & 89.9 & 90.2 & 48.5 & 83.3 & 65.9 & 78.4 & 44.5 & 62.4 & 65.5 & 43.2 & 54.3 \\
  
  & \ding{51} & \ding{55} & Strong & 96.4 & 93.0 & 94.7 & \bf 93.5 & 90.5 & 92.0 & \bf 78.9 & 49.8 & 64.4 & \bf 71.5 & 48.4 & 59.0 \\
  
  & \ding{51} & \ding{51} & Weak & 96.1 & 92.7 & 94.4 & 93.4 & 90.2 & 91.8 & 78.4 & 50.6 & 64.5 & 71.1 & 48.6 & 59.1 \\
  
  & \ding{51} & \ding{51} & Strong & \bf 96.6 & \bf 95.1 & \bf 95.8 & \bf 93.5 & \bf 93.3 & \bf 93.4 & 78.8 & \bf 52.0 & \bf 65.4 & \bf 71.5 & \bf 50.7 & \bf 61.1\\
 \bottomrule
  \end{tabular}
  \vspace{-2mm}
  \caption{Ablation study. Each core component of our method is removed in isolation. ``Concat'' stands for the concatenation of the logits before evaluating the softmax layer,  ``Over'' means overclustering and ``Aug'' is short for augmentation. We also report the performance of the full model to enable comparison. All the results reported in this table are measured on the test set and using the best head.}
  \label{tab:ablation}
\end{table*} 
\noindent\textbf{Implementation Details.} For a fair comparison with existing methods, we use a ResNet18~\cite{he2016deep} encoder for all datasets. The labeled head $h$ is an $l2$-normalized linear layer with $C^l$ output neurons, while the unlabeled head $g$ is composed of a projection head with 2048 hidden units and 256 output units, followed by a $l2$-normalized linear layer with $C^u$ output neurons. We pretrain our model for 200 epochs on the labeled dataset and then train for 200 epochs in the discovery phase on both labeled and unlabeled datasets. For both phases we use SGD with momentum as optimizer, with linear warm-up and cosine annealing ($lr_{base} = 0.1$, $lr_{min} = 0.001$), and weight decay $10^{-4}$. The batch size is always set to 512 for all experiments. Regarding the discovery phase, we use an overclustering factor $m = 3$ and a $n=4$ heads for both clustering and overclustering. The temperature parameter $\tau$ is set to $0.1$ for all softmax layers. 
For what concerns pseudo-labeling, we use the implementation of the Sinkhorn-Knopp algorithm~\cite{cuturi2013sinkhorn} provided by \cite{caron_swav} and we inherit all the hyperparameters from \cite{caron_swav}, \eg $n\_iter = 3$ and $\epsilon = 0.05$.

\noindent\textbf{Pretraining.}
In \cite{han2020automatically}, a three stage pipeline was proposed, where the network is first trained with self-supervision on the combination of labeled and unlabeled sets, then finetuned on the labeled set, and finally used to discover new classes. This complex procedure makes NCD cumbersome and costly. In addition, it is based on the assumption that the unlabeled data is available at training time, which might not always hold in real world scenarios (\textit{e.g.} online settings). For these reasons we decide not to use self-supervised pretraining in our method and show that it is enough to use a unified objective at discovery time in order to obtain the best performance. Nonetheless, for the sake of completeness, we examine the behavior of our model with self-supervised pretraining, finding no improvement with respect to simpler and more practical supervised pretraining. In our method, the labeled head $h$ is composed by $l2$-normalized prototypes and the last layer computes the cosine similarity between each prototype and the $l2$-normalized features $\zvect$. For consistency and performance, we also normalize such features and prototypes during pretraining.

\subsection{Ablation study}
In Tab.~\ref{tab:ablation} we report the results of an ablation study, obtained by removing each core component of our method in isolation, \textit{i.e.}, logit concatenation, overclustering, and augmentation.
For better inspecting the behavior of our model, we report clustering accuracy on both test subsets (labeled and unlabeled) as well as using the two proposed evaluation settings (task-aware and task-agnostic).

\noindent\textbf{Logit Concatenation.} As described in Sec.~\ref{sec:uno}, our main contribution is a unified objective for training on labeled and unlabeled data jointly. In other words, our model works by concatenating labeled and unlabeled logits in order to predict a posterior probability distribution over all classes. We experimentally demonstrate that this design choice is indeed crucial for the final performance of our method. The ablation shows that treating clustering and supervised learning with different objectives is highly suboptimal with respect to using our unified objective. In particular, the results point out two aspects. First, it is clear that using separated objectives yields greater interference between the two tasks. Importantly, this effect is also present when evaluating them separately (task-aware). Second, performance drops more dramatically when using the task-agnostic evaluation, especially on the labeled set. This latter result is justified by the fact that training without concatenation does not encourage the network to distinguish labeled from unlabeled samples.

\noindent\textbf{Overclustering.} In Sec.~\ref{sec:details} we described how we extract fine-grained clusters from the unlabeled data. This is known to significantly boost the quality of the representation~\cite{ji2019invariant,caron_swav}. We investigate this effect in our framework, finding evidence that overclustering can be effectively leveraged in NCD. Tab.~\ref{tab:ablation} shows that the performance on the unlabeled set is improved when the overclustering heads are used. Note that clustering heads are retained and used for evaluation, while overclustering heads are discarded at test time. Interestingly, the performance on the labeled classes does not benefit from fine-grained cluster extraction. This is reasonable because good representations of the labeled data are already learned using supervision. However, the overall accuracy is consistently higher when using overclustering, thus motivating our choice. 

\begin{table}[t]
  \centering
\resizebox{\columnwidth}{!}{
  \begin{tabularx}{1.3\columnwidth}{lcccc}
    \toprule
    Method & CIFAR10 & {CIFAR100-20} & CIFAR100-50 & ImageNet \\
    \midrule
    $k$-means~\cite{macqueen1967_kmeans} & 72.5$\pm$0.0 & 56.3$\pm$1.7 & 28.3$\pm$0.7 & 71.9 \\
    KCL~\cite{hsu2017learning} & 72.3$\pm$0.2 & 42.1$\pm$1.8 & - & 73.8\\
    MCL~\cite{hsu2019multi} & 70.9$\pm$0.1 & 21.5$\pm$2.3 & - & 74.4 \\
    DTC~\cite{han2019learning} & 88.7$\pm$0.3  & 67.3$\pm$1.2 & 35.9$\pm$1.0 & 78.3 \\
    RS~\cite{han2020automatically} & 90.4$\pm$0.5 & 73.2$\pm$2.1 & 39.2$\pm$2.3 & 82.5 \\
    RS+~\cite{han2020automatically} & 91.7$\pm$0.9 & 75.2$\pm$4.2 & 44.1$\pm$3.7 & 82.5 \\
    \midrule
    \bf UNO (avg) & \bf 96.1$\pm$0.5 & \bf 84.5$\pm$1.0 & \bf 52.8$\pm$1.4 & \bf 89.2 \\
    \bf UNO (best) & \bf 96.1$\pm$0.5 & \bf 85.0$\pm$0.6 & \bf 52.9$\pm$1.4 & \bf 90.6 \\
    \bottomrule
  \end{tabularx}
  }
  \vspace{-2mm}
  \caption{Comparison with state-of-the-art methods on CIFAR-10, CIFAR-100 and ImageNet for novel class discovery using task-aware evaluation protocol. Clustering accuracy is reported on the unlabeled set (training split). All methods except UNO initialize the encoder with self-supervised learning, except when evaluated on ImageNet. ``RS+'' is \cite{han2020automatically} with incremental classifier.}
  \label{tab:state-of-the-art}
\end{table}

\noindent\textbf{Data augmentation.} 
Very recently, the usage of strong data augmentation techniques has been thoroughly investigated in the context of self-supervised learning~\cite{simclr}. At the same time, unsupervised clustering techniques based on self-supervision have appeared in the literature~\cite{van2020scan,rebuffi2020lsd}. We follow these works and investigate the benefits of using SimCLR-like augmentations for NCD. First, we found that using very small crops does not improve the performance. Rather, it prevents the network from learning meaningful clusters. We believe this behavior is reasonable, because cropping occludes important parts of the image, making it hard for the network to produce meaningful predictions, which in turn decreases the quality of the pseudo-labels. Instead, using moderate random cropping (as found in~\cite{han2020automatically}) turned out to be the best choice in our experiments. Moreover, we discover that using strong color jittering and greyscale is beneficial for our method.

In Tab.~\ref{tab:ablation}, we evaluate two types of data augmentation:  \textbf{weak} (moderate random crop and random flip) versus \textbf{strong} (moderate random crop, flip, jittering, and greyscale). On ImageNet we also make use of Gaussian blur. From the results, it emerges that by using strong augmentations we consistently increase the accuracy of our method on both labeled and unlabeled sets. For a fair comparison, we also apply these strong transformations to RS~\cite{han2020automatically} without obtaining performance improvements.

\subsection{Comparison with the state-of-the-art}
We compare our approach with the current state-of-the-art in NCD: including KCL~\cite{hsu2017learning},  MCL~\cite{hsu2019multi},  DTC~\cite{han2019learning},  RS~\cite{han2020automatically} and RS+~\cite{han2020automatically} (``incremental classifier'' version of RS). In addition, we report the performance of k-means applied on top of pretrained features.

In Tab.~\ref{tab:state-of-the-art}, we focus on the unlabeled classes, reporting clustering accuracy on the training set (common practice in the literature~\cite{han2020automatically, han2019learning}). For all related methods we report results using self-supervised pretraining, as described in Sec.~\ref{sec:setup}, except for ImageNet, on which we only report results using supervised pretraining. Results using supervised pretrainig only for related methods are deferred to the supplementary material. Despite its simplicity and the lack of self-supervised pretraining, UNO considerably outperforms the state-of-the-art (RS+~\cite{han2020automatically}, which includes self-supervised pretraining), in some cases by $\approx$10\%. On CIFAR10 the clustering error is reduced to roughly half, coming very close to supervised accuracy. On ImageNet, UNO reaches over 90.0\% accuracy, a remarkable result given the complexity of the dataset. We believe such strong results validate our hypothesis that unifying clustering and supervised objectives is a more effective solution for NCD. The difference in performance between ``avg'' and ``best'' is negligible for simple datasets, while when it becomes larger for complex dataset with a higher number of classes. 
\begin{table}[t]
  \centering
  \resizebox{\columnwidth}{!}{
  \begin{tabular}{lccccccccc}
    \toprule
    \multirow{2}[3]{*}{Method}  &
    \multicolumn{3}{c}{CIFAR10} & \multicolumn{3}{c}{CIFAR100-20} & \multicolumn{3}{c}{CIFAR100-50}\\
    \cmidrule(lr){2-4} \cmidrule(lr){5-7} \cmidrule(lr){8-10}
     & Lab & Unlab & All & Lab & Unlab & All & Lab & Unlab & All \\
    \midrule
    KCL~\cite{hsu2017learning} & 79.4 & 60.1 & 69.8 & 23.4 & 29.4 & 24.6 & - & - & - \\
    MCL~\cite{hsu2019multi} & 81.4 & 64.8 & 73.1 & 18.2 & 18.0 & 18.2 & - & - & - \\
    DTC~\cite{han2019learning} & 58.7 & 78.6 & 68.7 & 47.6 & 49.1 & 47.9 & 30.2 & 34.7 & 32.5  \\
    RS+~\cite{han2020automatically} & 90.6 & 88.8 & 89.7 & 71.2 & 56.8 & 68.3 & 69.7 & 40.9 & 55.3 \\
    \midrule
    \bf UNO (avg)  & \bf 93.5 & \bf 93.3 & \bf 93.4 & \bf 73.2 & \bf 72.7 & \bf 73.1 & \bf 71.5 & \bf 50.6 & \bf 61.0  \\
    \bf UNO (best)  & \bf 93.5 & \bf 93.3 & \bf 93.4 & \bf 73.2 & \bf 73.1 & \bf 73.2 & \bf 71.5 & \bf 50.7 & \bf 61.1  \\

    \bottomrule
  \end{tabular}
  }
  \vspace{-2mm}
  \caption{Comparison with state-of-the-art methods on CIFAR-10 and CIFAR-100 on both labeled and unlabeled classes, using task-agnostic evaluation protocol. Accuracy and clustering accuracy are reported on the test set.}
  \label{tab:concat_eval}
\end{table}

We also thoroughly compare our method with the state-of-the-art in the task-agnostic evaluation setting on the test set. The results on CIFAR10 and CIFAR100-50 are shown in Tab.~\ref{tab:concat_eval}. For UNO, we report the average accuracy over the clustering heads. These results show that not only our method is better at clustering the unlabeled data, but it also outperforms the state-of-the-art on the labeled test set, showing that our unified objective favors better cooperation and less interference between labeled and unlabeled heads. Moreover, we notice that on CIFAR100-20 the relative improvement in clustering accuracy with respect to RS+~\cite{han2020automatically} is larger when using the task-agnostic evaluation with respect to the task-aware evaluation. This means that UNO also discriminates labeled from unlabeled classes better than the related methods.

\begin{figure}[t]
  \centering
  \includegraphics[height=0.32\columnwidth]{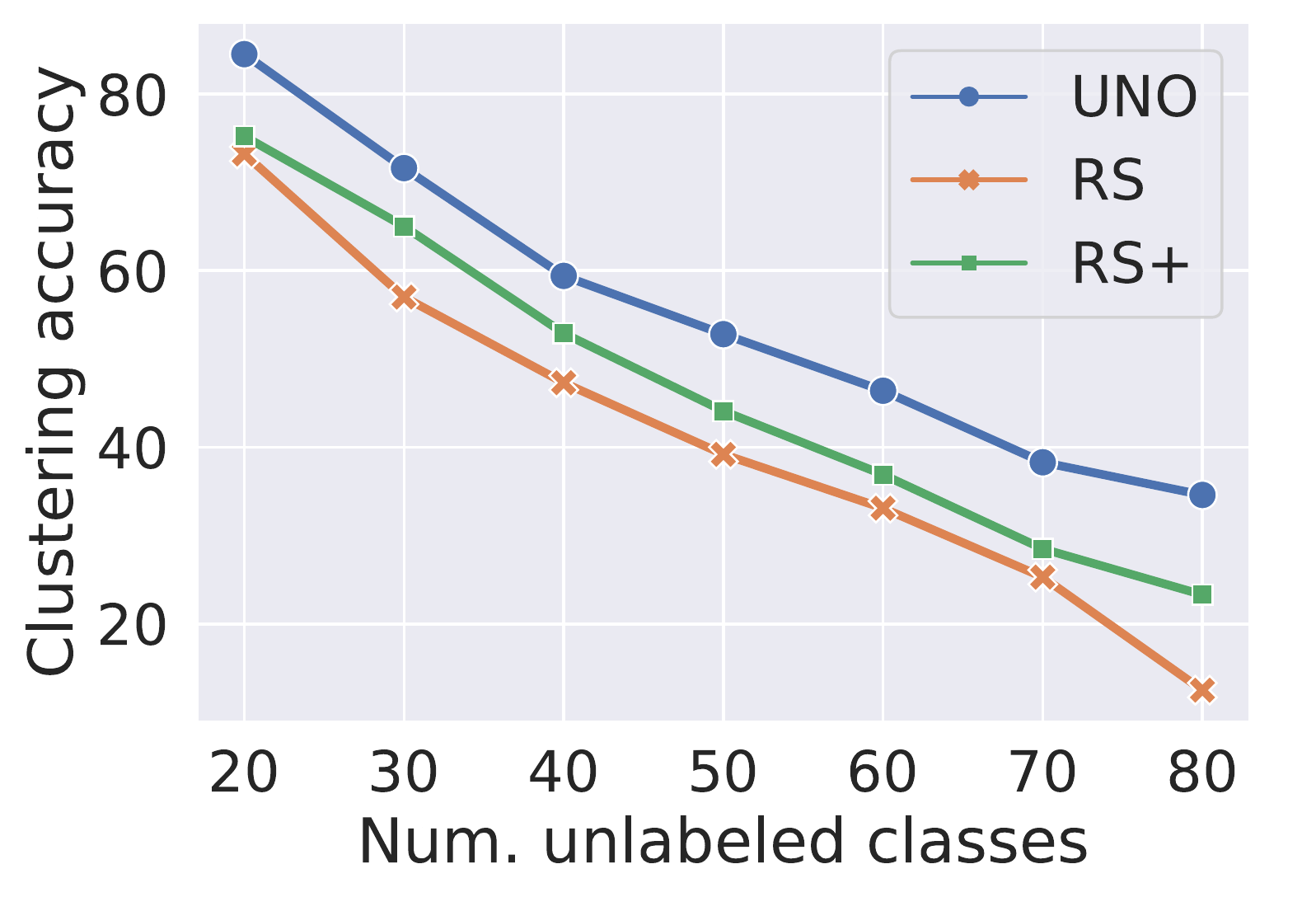}
  \includegraphics[height=0.32\columnwidth]{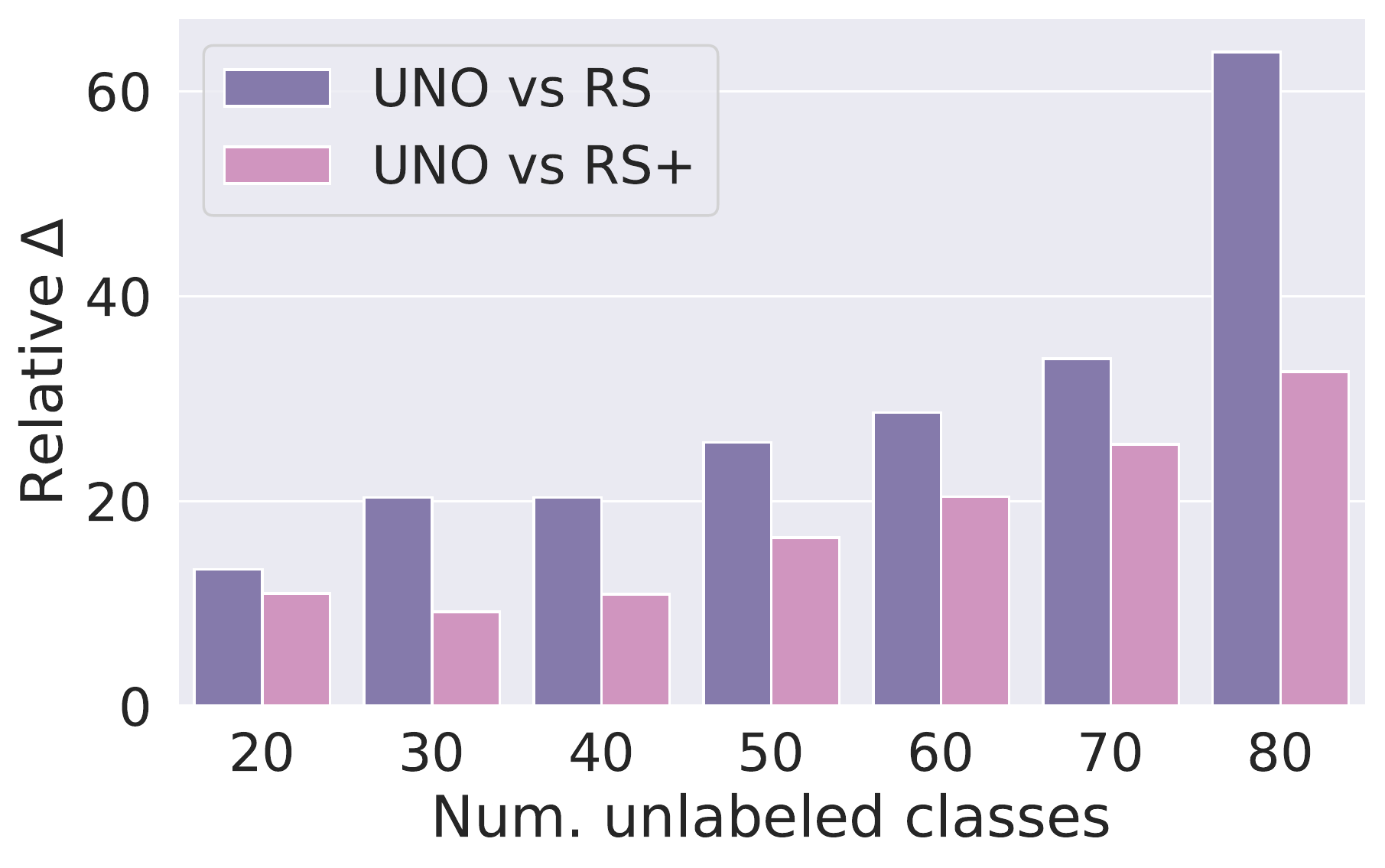}
    \vspace{-2mm}

  \caption{Clustering accuracy (left) and relative delta (right) with an increasing number of unlabeled classes. The relative delta is calculated as the margin between UNO and the related methods and normalized by the accuracy of UNO.}
  \label{fig:accu_num_unlab}
\end{figure}

Finally, in Fig.~\ref{fig:accu_num_unlab} we inspect the behavior of the best methods (Ours, RS and RS+) with an increasing number of unlabeled classes $C^u$. The task-aware clustering accuracy of all methods decreases as the task becomes increasingly difficult. This happens for two reasons: first, clustering becomes harder with more classes; second, the number of labeled classes $C^l$ decreases and in turn the quality of the representations learned with supervision is reduced. The latter is particularly disadvantageous for our model, since RS and RS+ are pretrained with self supervision on the union of the two datasets while ours only uses supervision on the labeled set. Despite this, our method always outperforms the other methods by large margins. Moreover, as shown in Fig.~\ref{fig:accu_num_unlab}  (right), the relative gap between our method and the others grows larger when $C^u$ increases, demonstrating the superiority of our objective even in complex scenarios.

\subsection{Qualitative results}

\begin{figure}[t]
  \centering
  \includegraphics[width=0.9\columnwidth]{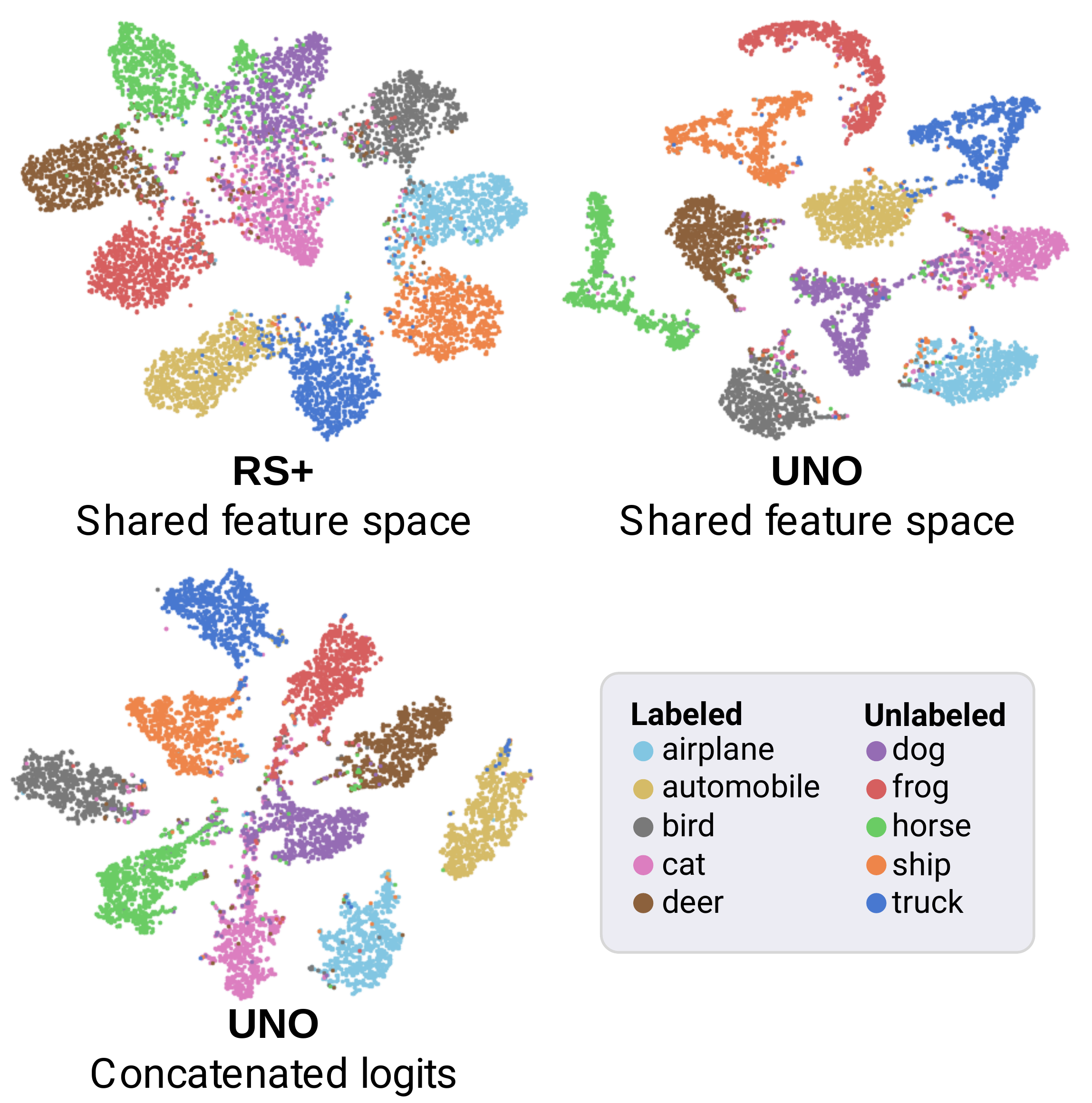}
  \vspace{-2mm}
  \caption{t-SNE visualization for all classes on CIFAR10. For both methods ``Shared feature space'' stands for the features after the last convolutional block.}
  \label{fig:features_all}
\end{figure}

In addition to quantitative results, we also report a qualitative analysis showing the feature space learned by our unified objective on CIFAR10. In Fig.~\ref{fig:features_all}, we visualize the shared feature space (after the last convolutional block) and the concatenated of the logits $\lvect$ of the heads $h$ and $g$. Since we have multiple unlabeled heads, for all data samples we concatenate the logits of all heads for visualization purposes. For the features, we run PCA~\cite{Jolliffe2011} to reduce their dimensionality. Finally we project the data in two dimensions using t-SNE~\cite{TSNE}. The same procedure is applied to the features produced by RS+~\cite{han2020automatically}.

From the plot, it is clear that our model produces feature representations for samples of the same class that are more tightly grouped with respect to RS+. At the same time, in RS+ several classes are entangled together (\eg cat, dog, horse), making it hard for a linear classifier to discriminate the samples. In accordance with our quantitative results, our method does a better job at separating the classes, both in the shared feature space and in the logits space.

Furthermore, we examine the influence that our architectural design choices have on the representations. It is clear from the plot that in the logits space samples are roughly uniformly distributed around the centroid of their class. On the other hand, in the shared feature space, labeled samples are still organized in disk-shaped groups, while unlabeled samples exhibit more irregular shapes. This is justified by the fact that the labeled head $h$ projects features linearly into logits, while the unlabeled head $g$ contains multiple layers and non-linearities. Moreover, unlabeled samples in the feature space seem to be organized in subgroups. This is probably due to the use of overclustering.

\section{Conclusions}
We presented a simple approach for discovering and learning novel classes in an unlabeled dataset, while leveraging good features extracted with supervision in a labeled dataset. Our method stands out from the literature for the fact that we use pseudo-labels in combination with ground truth labels in a UNified Objective (UNO) that enables better cooperation and less interference between supervised and unsupervised learning. Moreover, we also removed the need for costly self-supervised pretraining, making NCD more practical. We demonstrated the effectiveness of our proposed approach through extensive experiments and careful analysis. We found that UNO outperforms all related methods significantly, despite being conceptually simpler and easier to implement and train.

\noindent \small{\textbf{Acknowledgements.} 
This  work  is  supported  by the European Institute of Innovation \& Technology (EIT) and the H2020 EU project SPRING - Socially Pertinent Robots in Gerontological Healthcare. This work was carried out under the ``Vision and Learning joint Laboratory'' between FBK and UNITN.}

\renewcommand\thesection{\Alph{section}}
\renewcommand\thefigure{\Alph{figure}}
\renewcommand\thetable{\Alph{table}}
\setcounter{section}{0}
\setcounter{figure}{0}
\setcounter{table}{0}

\section*{Appendix}

\section{Comparison with the state-of-the-art without self-supervised pretraining}
In the main paper we presented a comparison with the state-of-the-art, using each method in its best possible configuration. For all competitors the best performance is reached by using self-supervised pretraining. This is disadvantageous for our method, since UNO only requires supervised pretraining. Also, using self-supervision makes all competing methods more computationally expensive. 

Hence, in Tab.~\textcolor{red}{1} we also report results without self-supervised pretraining, \ie with supervised pretraining only. In this comparison, all methods are trained using roughly the same amount of compute.  Clearly, all methods except UNO are negatively affected, \eg RS loses $\approx$$6$\%, DTC $\approx$$10$\% on CIFAR100-20. As a consequence, UNO outperforms the competitors even more significantly (\eg $6.7$\% and $17.6$\% on CIFAR10 and CIFAR100-20 respectively). This is a clear sign that, while UNO is able to learn powerful representations at discovery time, other methods need ad-hoc offline pretrainings that are often not possible in real world scenarios.

\begin{table}[h]
  \centering
  \begin{tabularx}{\columnwidth}{lcccc}
    \toprule
    Method & CIFAR10 & {CIFAR100-20} & ImageNet \\
    \midrule
    $k$-means~\cite{macqueen1967_kmeans} & 65.5$\pm$0.0\ & 56.6$\pm$1.6 & 71.9\\
    KCL~\cite{hsu2017learning} & 66.5$\pm$3.9 & 14.3$\pm$1.3 & 73.8 \\
    MCL~\cite{hsu2019multi} & 64.2$\pm$0.1 & 21.3$\pm$3.4 & 74.4 \\
    DTC~\cite{han2019learning}  & 87.5$\pm$0.3 & 56.7$\pm$1.2 & 78.3\\
    RS~\cite{han2020automatically}  & 89.4$\pm$1.4 & 67.4$\pm$2.0 & 82.5 \\
    \midrule
    \bf UNO (avg) & \bf 96.1$\pm$0.5 & \bf 84.5$\pm$1.0 & \bf 89.2 \\
    \bf UNO (best) & \bf 96.1$\pm$0.5 & \bf 85.0$\pm$0.6 & \bf 90.6 \\
    \bottomrule
  \end{tabularx}
  \caption{Comparison with state-of-the-art methods on CIFAR10, CIFAR100-20 and ImageNet for novel class discovery using task-aware evaluation protocol. Clustering accuracy is reported on the unlabeled set (training split). All methods initialize the encoder with supervised learning on the labeled set. ``RS+'' is with incremental classifier.}
   \label{tab:sota_supervised}

\end{table} 

\begin{figure*}[th!]
\begin{subfigure}{0.33\textwidth}
  \centering
  \includegraphics[width=.99\linewidth]{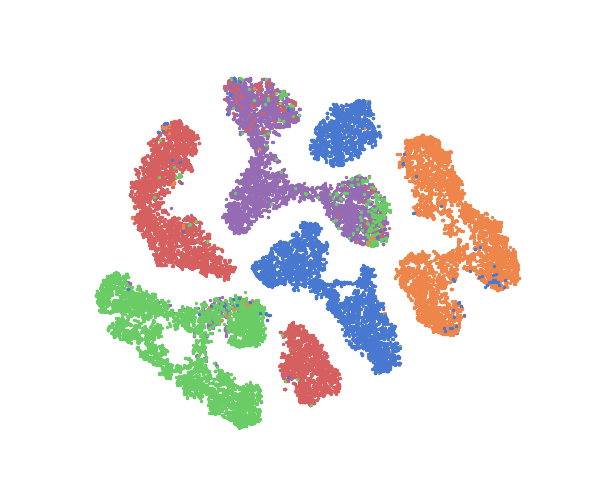}
  \caption{Shared feature space}
  \label{fig:feats_unlab_shared}
\end{subfigure}%
\begin{subfigure}{.33\textwidth}
  \centering
  \includegraphics[width=.99\linewidth]{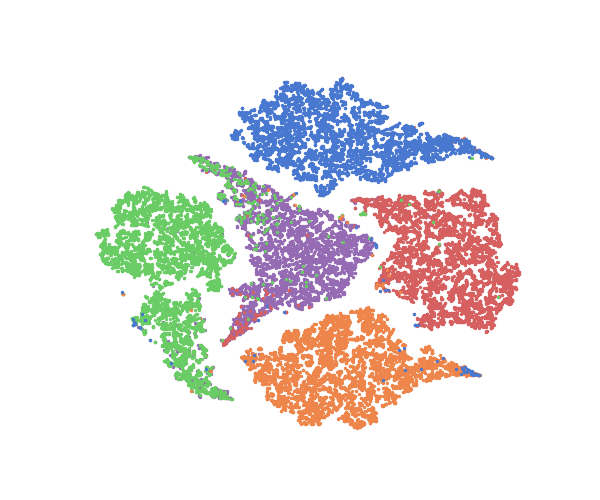}
  \caption{Clustering head logits}
  \label{fig:feats_unlab_cluster}
\end{subfigure}
\begin{subfigure}{.33\textwidth}
  \centering
  \includegraphics[width=.99\linewidth]{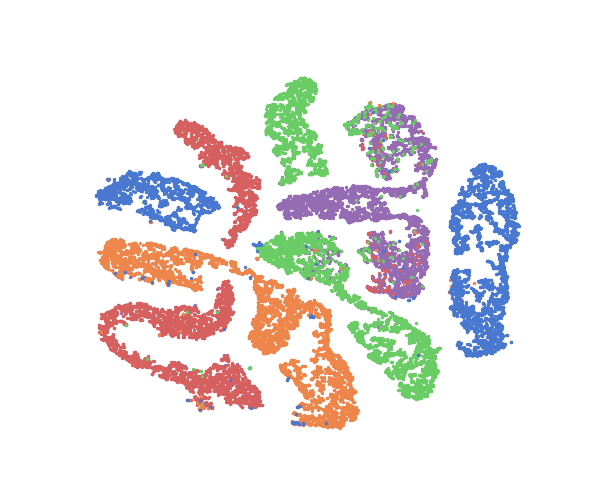}
  \caption{Overclustering head logits}
  \label{fig:feats_unlab_over}
\end{subfigure}
\caption{t-SNE visualization of unlabeled samples (training set) on CIFAR10.}
\label{fig:feats_unlab}
\end{figure*}

\section{Multi-view aggregation strategies}
In Sec.~\textcolor{red}{3} in the main manuscript, we described a way to generate pseudo-labels using multiple views. This corresponds to the \textbf{swapped prediction task} proposed in~\cite{caron_swav}. Nonetheless, other strategies can be employed. Instead of predicting the pseudo-label generated by another view, one can think of having a single pseudo-label that depends on all the views. This can be done in various ways. In the following we describe two aggregation strategies we investigated.

\noindent \textbf{Averaging pseudo-labels.} 
We generate pseudo-labels independently for each view using the Sinkhorn-Knopp algorithm~\cite{cuturi2013sinkhorn}. Then, we aggregate the pseudo labels by simply averaging the pseudo label over the views. In the case of two views this is equivalent to computing the following:
\begin{equation}
    \hat{\yvect} = \frac{\hat{\yvect}_1 + \hat{\yvect}_2}{2}.
\end{equation}
Subsequently, $\hat{\yvect}$ can be plugged in Eq.~(\textcolor{red}{2}) of the main paper to obtain the complete pseudo-label. 

This approach has advantages and disadvantages with respect to the swapped prediction task. For instance, an advantage is that the averaged pseudo-label is surely less noisy, since it depends on both views. However, averaging generates more entropic probability distributions, especially at the beginning, which slows down training. The parameters of the Sinkhorn-Knopp algorithm~\cite{cuturi2013sinkhorn} can be tuned to account for the smoothing introduced by averaging. However, these parameters depend on the number of views, which makes this strategy impractical. Nonetheless, as shown in Tab.~\ref{tab:aggregation}, this aggregation strategy produces very similar results to the swapped prediction task when using two views.

\noindent \textbf{Averaging logits.}
Similarly, another solution to generate aggregated pseudo-labels is to first average the logits:
\begin{equation}
    \lvect_g = \frac{\lvect_g^1 + \lvect_g^2}{2},
\end{equation}
then generate $\hat{\yvect}$ from $\lvect_g$, and finally plug Eq.~(\textcolor{red}{2}) of the main paper. This solution is particularly advantageous in terms of computation, since it requires us to run Sinkhorn-Knopp~\cite{cuturi2013sinkhorn} only once, instead of twice (in the case of two views). However, unfortunately, this strategy does not produce results that are as good as using the swapped prediction task. (see Tab.~\ref{tab:aggregation})

\begin{table}[t]
  \centering
  \begin{tabular}{lcccc}
    \toprule
    Method & Aggregation & CIFAR10 & CIFAR100-50 \\
    \midrule
    \multirow{3}{*}{UNO}
    & logits & 94.4 & 52.2 \\
    & pseudo-labels & \bf 96.1 & 52.5 \\
    & swap & \bf 96.1 & \bf 52.9 \\
    \bottomrule
  \end{tabular}
  \caption{Comparison of multi-view aggregation strategies. ``logits'' stands for \textbf{averaging logits}, ``Pseudo-labels'' for \textbf{averaging pseudo-labels} and ``swap'' for the \textbf{swapped prediction task}. We report the clustering accuracy of the best head on the training set.}
  \label{tab:aggregation}
\end{table} 

\section{More qualitative results for UNO}
In this section, we show additional qualitative results and analysis of the feature space induced by our Unified Objective (UNO). In the main paper we reported a visual comparison of the features extracted by UNO w.r.t RS, showing a clear advantage of our method. Here, we dig deeper and investigate how clustering and overclustering heads project the features. For visualization purposes, we concatenate the logits of the multiple heads we use for clustering and overclustering respectively.

In Fig.~\ref{fig:feats_unlab_shared} the reader can appreciate how unlabeled classes are organized in subgroups in the shared feature space. This is due to the use of overclustering. Nonetheless, these subgroups are tightly clustered and can be easily separated from samples of other classes. Indeed, as shown in Fig.~\ref{fig:feats_unlab_cluster}, the non linear projection head we make use of can correctly group most of the samples. Interestingly, the overclustering heads project the features in a way that increases the separation of the subgroups (see Fig.~\ref{fig:feats_unlab_over}), also sometimes stretching them in an attempt to minimize the loss.

\begin{table}[t]
  \centering
  \begin{tabular}{lc}
    \toprule
    Method & CIFAR100-20 \\
    \midrule
    DTC~\cite{han2019learning} & 64.3 \\
    RS~\cite{han2020automatically}  & 70.5 \\
    RS+~\cite{han2020automatically}  & 71.2 \\
    \midrule
    \bf UNO (avg) & \bf 74.7 \\
    \bf UNO (best) & \bf 75.1 \\
    \bottomrule
  \end{tabular}
  \caption{Comparison with state-of-the-art methods on CIFAR100-20 and ImageNet for novel class discovery with unknown number of classes $C^u$, using task-aware evaluation protocol. Clustering accuracy is reported on the unlabeled set (training split). All methods except UNO initialize the encoder with self-supervised learning.}
  \label{tab:unknown_number}
\end{table}

\section{Unknown number of clusters}
All the results we showed so far assumed the knowledge of the number of classes $C^u$ contained in the unlabeled set. However, in practical scenarios, it is unlikely to dispose of that information. While many previous works investigated the problem of estimating the number of clusters given a set of unlabeled data~\cite{macqueen1967_kmeans}, in the context of NCD an effective approach was proposed in~\cite{han2019learning}. This method consists in holding out a probe subset from the labeled set, and then running a constrained (semi-supervised) $k$-means routine on the union of the probe subset and unlabeled set. Subsequently, the optimal number of clusters $k$ is estimated using clustering quality indices on both subsets.

To investigate the applicability of our Unified Objective (UNO) in practical scenarios where $k$ is not available, we estimate the number of clusters on CIFAR100-20, using the aforementioned approach described in~\cite{han2019learning}. We use 60 classes for feature pretraining, 20 classes in the probe subset and 20 classes in the unlabeled set. In this way, we obtain a reasonable estimation, $k=23$ classes. We then rerun UNO and the competing methods using this estimation. The results are shown in Tab~\ref{tab:unknown_number}. We find that our method still outperforms the state-of-the-art considerably.

\begin{table}[t]
  \centering
  \begin{tabularx}{\columnwidth}{lcccc}
    \toprule
    Method & CIFAR10 & {CIFAR100-20} & ImageNet \\
    \midrule
    Jia \etal~\cite{jia21joint} &  93.4$\pm$0.6 & 76.4$\pm$2.8 &  86.7 \\
    OpenMix~\cite{zhong2021openmix} &  95.3 &  - &  85.7 \\
    NCL~\cite{Zhong_2021_CVPR} &  93.4$\pm$0.5 &  \bf 86.6$\pm$0.4 &  \bf 90.7 \\
    \midrule
    \bf UNO (avg) & \bf 96.1$\pm$0.5 & 84.5$\pm$1.0 &  89.2 \\
    \bf UNO (best) & \bf 96.1$\pm$0.5 & 85.0$\pm$0.6 &  90.6 \\
    \bottomrule
  \end{tabularx}
  \caption{Comparison with concurrent methods on CIFAR10, CIFAR100-20 and ImageNet for novel class discovery using task-aware evaluation protocol. Clustering accuracy is reported on the unlabeled set (training split).}
  \label{tab:concurrent}
\end{table} 

\section{Comparison with concurrent works}
In this section we compare the performance obtained with UNO to the following concurrent works: OpenMix~\cite{zhong2021openmix}, NCL~\cite{Zhong_2021_CVPR} and Jia \etal~\cite{jia21joint}. The results can be found in Tab.~\ref{tab:concurrent}. Despite being much simpler than all the concurrent related methods, UNO achieves better or comparable performance. In particular, UNO still achieves state-of-the-art results on CIFAR10 while it is slightly outperformed by NCL~\cite{Zhong_2021_CVPR} on CIFAR100-20 and ImageNet.
\clearpage

{\small
\bibliographystyle{ieee_fullname}
\bibliography{bibliography}
}

\end{document}